\documentclass{article}

\usepackage{PRIMEarxiv}

\usepackage[utf8]{inputenc}
\usepackage[T1]{fontenc}
\usepackage[hypertexnames=false,hidelinks]{hyperref}
\usepackage{url}
\usepackage{booktabs}
\usepackage{amsmath}
\usepackage{amsfonts}
\usepackage{multirow}
\usepackage{float}
\usepackage[numbers]{natbib}
\usepackage{microtype}
\usepackage{fancyhdr}
\usepackage{graphicx}

\title{Attention Expansion:\\
Enhancing Keyphrase Extraction from Long Documents\\
with Attention-Augmented Contextualized Embeddings}

\author{\normalfont
Roberto Mart{\'i}nez-Cruz$^{1,2,*}$, Alvaro J. L{\'o}pez-L{\'o}pez$^{1}$, and Jos{\'e} Portela$^{1}$\\[3pt]
\footnotesize $^{1}$Institute for Research in Technology, ICAI School of Engineering,\\
\footnotesize Comillas Pontifical University, Madrid, Spain.\\[3pt]
\footnotesize $^{2}$DD-AIM, New York, USA.\\[3pt]
\footnotesize $^{*}$Corresponding author: \texttt{rmcruz@comillas.edu}\\
\footnotesize Contributing authors: \texttt{allopez@comillas.edu}; \texttt{jportela@comillas.edu}
}

\begin{document}
\maketitle

\begin{abstract}
Pre-trained language models (PLMs) have achieved strong performance in keyphrase extraction (KPE), largely due to their ability to generate rich contextualized representations. However, long-document KPE remains challenging because salient keyphrase evidence may be scattered across distant document sections that cannot be jointly captured within the limited context window of most PLMs. Although long-context large language models (LLMs) can process broader textual contexts, their computational cost limits their practicality for efficient and high-throughput KPE. To overcome this limitation, we propose an attention expansion mechanism that augments PLM token representations with information from surrounding out-of-context chunks using pre-trained word embeddings. The proposed mechanism expands the effective contextual scope of PLM-based KPE models without requiring full-document attention or expensive LLM-based inference. We evaluate our approach across five PLM backbones, including general-purpose, scientific, task-specific, and long-context encoders, using two training regimes and five benchmark corpora from scientific and news domains. Experimental results demonstrate that attention expansion consistently enhances KPE performance across all evaluation settings, outperforming state-of-the-art models and yielding notable improvements in F1 score. The improvements extend to domain-specific, task-specialized, and native long-context models, showing that the proposed mechanism provides complementary information rather than merely compensating for limited input length. These results establish attention expansion as an efficient and effective strategy for long-document KPE.
\end{abstract}

\section{Introduction}

Keyphrase extraction (KPE) aims to identify the phrases that best capture the main themes of a document. As a compact form of document understanding, KPE supports a wide range of downstream applications, including document classification \citep{hulth2006study}, clustering \citep{hammouda2005corephrase}, summarization \citep{summarization-example-1, summarization-example-2}, indexing \citep{ie-example}, query expansion \citep{query-expansion-example}, and interactive document retrieval \citep{jones1999phrasier}. Often described as keyphrasification \citep{keyphrasification}, the task can be viewed as an extreme form of extractive summarization in which a small set of phrases must preserve the document's essential content.

The need for robust KPE is particularly acute for long documents. In domains such as science, law, medicine, and finance, important information is often distributed across multi-page texts in which the significance of a term emerges only when related evidence from distant sections is jointly considered. Many existing KPE pipelines partly avoid this difficulty by relying on short summaries or abstracts, especially in scientific corpora. While effective in some benchmark settings, this assumption is restrictive in real-world scenarios, where summaries may be unavailable, incomplete, or insufficiently informative. Moreover, important extractive keyphrases can be absent from abstracts even when they are central to the full document. As a result, methods that perform well on short inputs do not necessarily transfer well to long-document KPE.

KPE methods are commonly divided into unsupervised and supervised approaches. Unsupervised methods typically rank candidate phrases using statistical or graph-based criteria \citep{graph-based-ke, boudin2013comparison}. Supervised methods instead formulate KPE as a token classification problem, representing tokens through handcrafted features \citep{hasan2014automatic}, static word embeddings such as word2vec and GloVe \citep{word2vec, glove}, or contextualized representations from pre-trained language models (PLMs) such as BERT \citep{bert, sahrawat2020keyphrase}. Among these alternatives, PLM-based methods have become especially effective because they provide context-sensitive token representations that can be directly fine-tuned for sequence tagging.

Despite their success, PLM-based KPE models remain constrained by limited context length. In the standard long-document setting, the input is split into windows that fit the model's maximum sequence length, and each window is processed independently. This strategy preserves local contextualization, but prevents the model from directly representing dependencies that span multiple windows. Such long-range dependencies are often crucial for identifying keyphrases in long texts, where topical salience may depend on repeated mentions, delayed definitions, or relationships between concepts introduced far apart in the document. Current state-of-the-art KPE systems, whether based on fine-tuned PLMs or on architectures that consume PLM embeddings \citep{kbir, sahrawat2020keyphrase, park2020scientific}, therefore inherit a fundamental limitation: they benefit from contextualized representations, yet remain bounded by the context window of the underlying model.

Architectural extensions of the transformer such as Longformer \citep{longformer} and BigBird \citep{bigbird}, and more recent encoder-only models such as ModernBERT \citep{modernbert}, push the native context length further, while long-context LLMs extend it further still. These developments confirm the value of broader context for language understanding, but broader context is not free: it increases computational cost, memory usage, and deployment complexity. In the high-throughput regimes in which KPE is typically deployed (large-scale indexing, retrieval-augmented generation, or document pre-processing for downstream reasoning), replacing a compact, specialized model with a long-context LLM is rarely the most efficient option. Specialized KPE models therefore retain a clear practical role, consistent with the broader argument, in Kahneman's fast-and-slow framing \citep{kahneman2011}, that fast and efficient task-specific components should complement, rather than be replaced by, larger generic reasoners. Recent philosophical analyses make a compatible distinction, arguing that the intelligent behaviour of AI systems can be explained without attributing human-like \emph{understanding}, as a mechanistic form of task-solving \citep{garrido_blanco_2024}; this reinforces the practical motivation for improving the representations available to compact specialized models rather than treating larger generic models as the only route to better KPE. The question we address is how to give such specialized models access to long-document evidence without paying the cost of full long-context attention or LLM-based inference.

To this end, we propose an \emph{attention expansion} mechanism for PLM-based KPE. The mechanism augments the standard contextualized representation produced by the PLM with information extracted from out-of-context document chunks. Each surrounding chunk is represented by a sequence of pre-trained word embeddings (PWE), which provide a compact lexical summary of regions of the document that fall outside the PLM's current window. A cross-attention layer then lets each in-context PLM token query these out-of-context PWE representations, retrieving and aggregating evidence that is otherwise unreachable. The resulting token representations combine the rich local contextualization of the PLM with broader document-level evidence drawn from neighbouring chunks, producing enriched embeddings that are then fed to the sequence-tagging classifier.

This design expands the effective context of the PLM without expanding the PLM itself. Because the surrounding chunks are encoded only with pre-trained word embeddings rather than with a second forward pass through the transformer, attention expansion avoids the quadratic cost of full long-context attention and the inference cost of an LLM call. Parameter growth is bounded by the static embedding dimension, the additional computation per token is linear in the number of out-of-context positions attended to, and the mechanism plugs into any encoder backbone that exposes a hidden state at the token level. As a result, attention expansion is well suited to settings where computational efficiency is critical, and it remains useful even for encoders that already support longer inputs: by surfacing complementary lexical evidence beyond what the PLM has been pre-trained to attend to, the mechanism continues to improve representations of long-context models such as ModernBERT.

Attention expansion is complementary to existing approaches that incorporate broader document information into KPE models. Graph-based document representations \citep{gnn-kpe} contribute a global structural signal, while recent long-document KPE methods such as LongKey \citep{longkey} and MAPEX \citep{mapex} aggregate evidence through chunk-wise pooling or multi-agent pipelines. These approaches typically combine signals \emph{outside} the PLM, after the contextualized embeddings have already been produced. By contrast, attention expansion introduces an explicit attention-based bridge between in-context PLM embeddings and out-of-context lexical representations, integrating long-range evidence \emph{inside} the token representation itself. The two families of methods can therefore be combined.

The main contributions of this paper are as follows:
\begin{itemize}
\item We introduce an attention expansion mechanism that augments PLM-based KPE with cross-attention from in-context token representations to pre-trained word embeddings of the surrounding out-of-context chunks, expanding the effective context of the model without invoking full long-context attention or LLM-based inference.
\item We show that attention expansion enriches PLM token representations in a parameter-light and computationally efficient way, with parameter growth bounded by the static embedding dimension and no additional full-attention passes over the input, making it suitable for high-throughput KPE pipelines.
\item We provide an extensive empirical evaluation across five encoder backbones (DistilBERT, SciBERT, KBIR, DeBERTa-v3, and ModernBERT), two training regimes (SemEval-2010 and LDKP3K), and five benchmark corpora from the scientific and news domains. Attention expansion consistently improves KPE performance and outperforms strong baselines, and its benefits extend to domain-specialized and native long-context encoders, indicating that the mechanism provides complementary information rather than merely compensating for limited input length.
\end{itemize}

\section{Related Work}
\label{related}

\subsection{Keyphrase Extraction Methods}
Keyphrase extraction (KPE) has traditionally been studied through unsupervised ranking and supervised prediction. Unsupervised systems first identify candidate phrases, commonly using lexical or part-of-speech patterns, and then rank candidates according to statistical, positional, topical, or graph-based salience \citep{hasan2014automatic}. Graph-ranking methods are particularly influential: TextRank ranks words in a co-occurrence graph using a PageRank-inspired procedure \citep{textrank}, while TopicRank clusters candidates into topics before ranking topic representatives \citep{topicrank}. Subsequent analyses have examined how centrality choices affect graph-based extraction \citep{boudin2013comparison}. Distributed representations later enriched this family of methods by measuring semantic similarity between phrases and their documents or themes \citep{wang2014corpus, mahata2018key2vec, mahata2018theme, bennani2018simple}. More recently, contextual embeddings and prompting have supported unsupervised ranking without task-specific labels: PatternRank combines part-of-speech patterns with PLM representations \citep{patternrank2022}, and PromptRank uses prompts to elicit phrase relevance from a pretrained model \citep{kong2023promptrank}. These methods remain attractive when annotation is unavailable or domain transfer is required.

When labeled data are available, KPE can instead be treated as span identification or sequence labeling. Earlier supervised approaches relied on handcrafted lexical, syntactic, and positional features; modeling token sequences with a conditional random field made it possible to exploit relationships between adjacent labels rather than classifying candidates independently \citep{Gollapalli_Li_Yang_2017}. Neural variants replaced much feature engineering with pretrained word embeddings and recurrent contextualization. For scientific documents, BiLSTM--CRF models using distributed word representations showed the benefit of jointly capturing contextual evidence and consistent phrase boundaries \citep{alzaidy_2019, patel2019exploring}. This line is closely aligned with extractive settings such as the present paper, where a BIO tagger predicts phrases that occur in the source document.

Transformers and pretrained language models (PLMs) substantially strengthened sequence-labeling KPE by producing token representations conditioned on surrounding text. Transformer taggers such as TransKP and TNT-KID developed task-oriented architectures for keyword identification \citep{transkp, tnt-kid}. Sahrawat et al.\ showed that contextualized PLM embeddings combined with sequence-labeling layers improve scientific KPE, with domain-adapted representations particularly useful for scientific language \citep{sahrawat2020keyphrase}. SciBERT similarly illustrates the value of pretraining on scientific corpora \citep{scibert}, while intermediate-task transfer has been studied for scientific keyphrase identification and classification \citep{park2020scientific}. KBIR goes further by introducing pretraining objectives that explicitly model keyphrase boundaries and replacements, yielding a task-specialized encoder for KPE \citep{kbir}. Taken together, these results establish contextual token representation as a central ingredient of strong supervised KPE systems. However, the representations produced by standard encoder models are only contextual within their maximum input window.

\subsection{Keyphrase Extraction from Long Documents}
This context limitation is especially consequential for full documents. Much KPE research has operated on titles or abstracts, where keyphrases and their supporting evidence are concentrated within short inputs. In scientific papers, technical reports, and news collections, by contrast, a salient phrase may be introduced in one section, elaborated in another, and repeated or contextualized much later. The LDKP datasets were introduced specifically to support evaluation on long scientific documents and demonstrate that methods designed for short inputs face a substantially different extraction setting \citep{mahata2022ldkp}. Related work on keyphrase generation beyond titles and abstracts likewise shows that restricting input to summaries omits useful document evidence \citep{garg2022keyphrase}.

A general response to long inputs is to enlarge the transformer's context. Longformer uses local sliding-window attention with selected global attention positions, and BigBird uses sparse attention patterns to reduce the cost of representing longer sequences \citep{longformer, bigbird}. ModernBERT provides a more recent encoder-only alternative with native 8,192-token support and engineering improvements intended for efficient long-context fine-tuning and inference \citep{modernbert}. These architectures make more text directly accessible than conventional 512-token encoders. Nevertheless, long-document KPE requires not only accessing additional tokens but also forming token or candidate representations that reflect evidence dispersed throughout a document; even longer native windows may be exceeded or may devote computation to text that is weakly relevant to extraction.

KPE-specific approaches therefore augment or restructure context rather than relying only on a larger backbone. Do\v{c}ekal and Smr\v{z} chunk lan long documents while preserving a global query that indicates document topic, showing that targeted global context can be more useful than simply increasing the local input span \citep{query-based-kpe}. LongKey performs chunk-wise encoding and consolidates repeated candidate occurrences through max-pooled candidate representations, explicitly reconnecting evidence observed in different segments \citep{longkey}. Unsupervised long-document work also emphasizes global candidate scoring: UFORank combines topic importance, positional evidence, and phrase-to-topic similarity in an embedding-based ranking framework \citep{uforank}. A complementary supervised direction augments PLM representations with document-level structural signals; graph-enhanced sequence tagging uses embeddings of document co-occurrence graphs to provide information about relationships extending beyond local PLM context \citep{gnn-kpe}. Such methods indicate that extending effective context is not synonymous with applying full self-attention to the entire document: document-level information can be summarized, pooled, queried, or represented structurally.

\subsection{LLM-Based Methods and the Remaining Gap}
Keyphrase generation (KPG) broadens the task by allowing absent phrases that do not occur verbatim in the document, whereas the BIO sequence-labeling setting addressed here is extractive and aims to identify source spans. Encoder-decoder and decoder-only generative models are therefore useful comparators, particularly because instruction-tuned large language models (LLMs) can process broader contexts and adapt through prompting. A benchmarking study of ChatGPT for KPG reports strong results across document lengths and domains, including full-length inputs \citep{martinezcruz2023chatgpt}. More recent independent work examines zero-shot extractive prompting directly: Kang and Shin evaluate multiple prompting strategies with instruction-tuned LLMs and report competitive unsupervised KPE results across six benchmarks \citep{kang-shin-2025-zero-shot}. MAPEX further structures LLM-based extraction as a multi-agent pipeline whose processing path depends on document length \citep{mapex}. LongDocRank combines LLM-generated candidates with graph-based reranking intended to recover document-level salience in long inputs \citep{longdocrank}.

These approaches make clear that LLMs and long-context architectures are relevant to long-document keyphrase prediction. They also represent a different operating point from a fine-tuned extractive encoder. Generative inference must produce and often post-process a set of strings rather than classify document tokens, and LLM pipelines may require repeated prompts, candidate generation, aggregation, or reranking. Their broader context and transfer capacity can be valuable when absent phrases or zero-shot behavior are central, but their computational demands are less suitable for high-throughput extractive indexing. Moreover, longer accessible context does not by itself ensure that the evidence governing phrase salience is effectively incorporated into each token decision.

The remaining gap is thus an efficient way to enrich supervised token-level KPE with distributed document evidence. PLM-based taggers provide strong local contextual representations, but processing long documents in independent windows fragments evidence outside each window. Native long-context encoders alleviate this boundary at increased computational cost and still have finite receptive ranges, while LLM-based extraction or generation trades compact token classification for substantially heavier inference. The present work addresses this gap by allowing in-context PLM token representations to attend to out-of-context static word embeddings from surrounding chunks. In this way, it seeks to retain the efficiency and controlled extractive output of sequence tagging while incorporating information that conventional windowed PLM encoding cannot directly represent.

\section{Methods}
\label{methodology}

In this section we describe the proposed attention expansion mechanism. We first frame KPE as a sequence tagging task, then recall the conventional sliding-window encoding used by PLMs on long documents, and finally introduce the attention expansion mechanism that augments PLM token representations with information from out-of-context document chunks.

\subsection{Problem Formulation}
\label{problem-formulation}
Let $d = \{w_1, w_2, \ldots, w_n\}$ denote an input document of $n$ words. The objective is to assign to each word $w_t$ one of three labels $Y = \{K_{B}, K_{I}, O\}$, where $K_{B}$ marks the first word of a keyphrase, $K_{I}$ marks a word that continues an already-started keyphrase, and $O$ indicates that $w_t$ does not belong to any keyphrase. This is the standard BIO tagging scheme for sequence labelling, and reduces KPE to a token-level classification problem in which adjacent $K_B / K_I$ labels jointly delimit a keyphrase span.

\subsection{Encoding Long Documents with PLMs}
\label{window-embedding}
PLM-based sequence taggers operate on a maximum input length, denoted $L = $ \textit{max\_length}, that is set by the model's positional embeddings. When the document length exceeds $L$, we follow the standard sliding-window approach and partition $d$ into a sequence of non-overlapping chunks $c_1, c_2, \ldots, c_m$, each of length at most $L$ tokens, padded to $L$ where needed. Each chunk is processed independently by the PLM, producing a sequence of contextualised token representations
$
H_i = \mathrm{PLM}(c_i) \in \mathbb{R}^{L \times d_{\mathrm{PLM}}}.
$
Word-level BIO labels are aligned to the sub-token sequence produced by the tokenizer by assigning the word's label to its first sub-token and a masking value to all remaining sub-tokens; special tokens (e.g.\ \texttt{[CLS]}, \texttt{[SEP]}, padding) receive the same masking value. During training, sub-tokens marked with the masking value are ignored by the cross-entropy loss, so the optimiser only updates the representation of word-initial positions. At evaluation time, the per-chunk predictions are concatenated back to a single sequence per document using a per-chunk document index, masked positions are dropped, and the resulting BIO sequence is scored at the entity level.

\begin{figure}[H]
    \centering
    \includegraphics[width=0.72\linewidth]{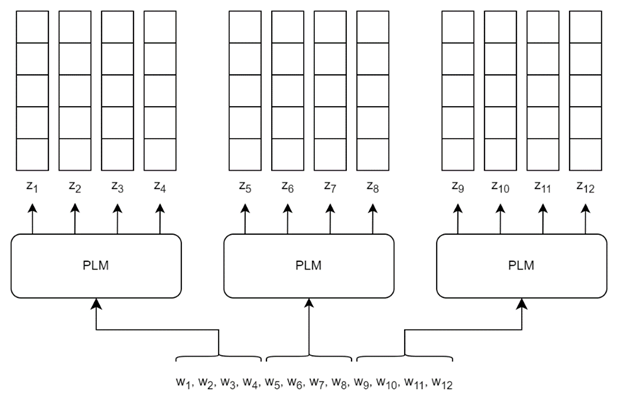}
    \caption{Schematic example of sliding-window embedding computation for long documents. The PLM sees only one chunk at a time, so token representations within a chunk cannot directly attend to words from other chunks.}
    \label{fig:long-plm-embedding}
\end{figure}

Under this scheme, the contextual information available to the encoder is restricted to the words contained in the same chunk: words separated by a chunk boundary cannot directly attend to one another, even when they are semantically related and central to the same keyphrase. Figure~\ref{fig:long-plm-embedding} illustrates this constraint on a 12-word document with $L = 4$.

\subsection{Attention Expansion Mechanism}
\label{attn-exp-embedding}
To overcome the per-chunk locality of the PLM, we introduce an attention expansion mechanism that lets each in-context token attend to representations of the \emph{surrounding} chunks without re-encoding them with the PLM. We use a Pre-trained Word Embedding (PWE) model as a lightweight encoder of the out-of-context regions, and a single cross-attention head to integrate that signal into the PLM token representations.

\paragraph{Context window.}
For a given chunk $c_i$ of length $L$, let $w$ denote the context window size (the total number of surrounding chunks attended to). We build the out-of-context sequence
\[
C_i = \left[\, c_{i-\tfrac{w}{2}}, \ldots, c_{i-1},\; c_{i+1}, \ldots, c_{i+\tfrac{w}{2}} \,\right],
\]
which concatenates the $w/2$ chunks immediately before and the $w/2$ chunks immediately after $c_i$. The chunk $c_i$ itself is \emph{not} included in $C_i$, since its content is already encoded by the PLM. When $c_i$ is close to a document boundary, the missing chunks are replaced by padding so that $|C_i| = w \cdot L$ tokens regardless of position.

\paragraph{Cross-attention.}
Let $E_{\mathrm{PWE}} : \mathbb{N} \to \mathbb{R}^{d_{\mathrm{PWE}}}$ denote a PWE function that maps a token id to a static word vector. The PLM produces
$
H_i \in \mathbb{R}^{L \times d_{\mathrm{PLM}}}
$
for the in-context chunk, while the surrounding chunks are summarised as
\[
S_i = E_{\mathrm{PWE}}(C_i) \;\in\; \mathbb{R}^{w L \times d_{\mathrm{PWE}}}.
\]
Queries are obtained by linearly projecting the PLM hidden states, and keys and values are obtained by linearly projecting the PWE representations. A single cross-attention head computes
\begin{equation}
A_i \;=\; \mathrm{softmax}\!\left( \frac{(H_i W^Q)\,(S_i W^K)^{\top}}{\sqrt{d_{\mathrm{PWE}}}} \right) (S_i W^K),
\label{eq:single-head-ae}
\end{equation}
where $W^Q \in \mathbb{R}^{d_{\mathrm{PLM}} \times d_{\mathrm{PWE}}}$ and $W^K \in \mathbb{R}^{d_{\mathrm{PWE}} \times d_{\mathrm{PWE}}}$. We deliberately \emph{tie} the key and value projections: the same matrix $W^K$ produces both the keys and the values. This design choice halves the number of attention parameters at no observed cost in downstream F1 and was retained from the initial implementation after preliminary experiments with untied projections showed no consistent improvement.

\begin{figure}[H]
    \centering
    \includegraphics[height=0.55\textheight,keepaspectratio]{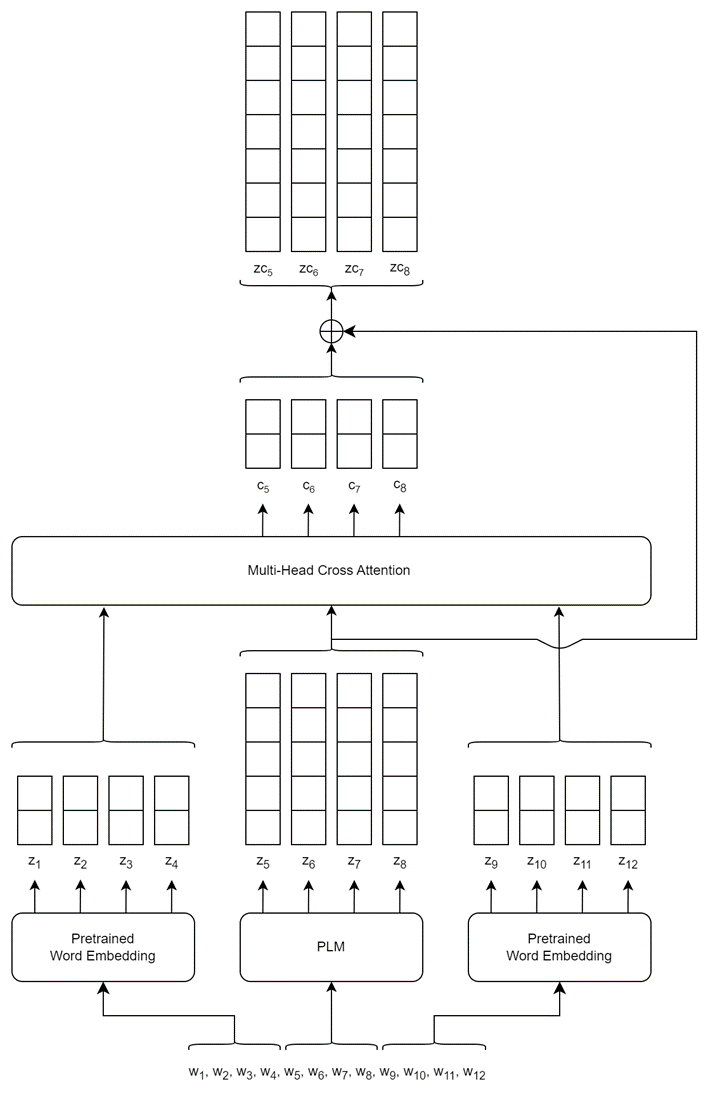}
    \caption{Schematic depiction of the attention expansion mechanism. The PLM contextualises a single chunk, while PWE provides lightweight representations for the $w$ surrounding chunks (chunk $c_i$ itself is excluded from the context). A cross-attention layer aggregates the surrounding evidence into a vector $A_i$, which is concatenated to the PLM hidden state before the BIO classification head.}
    \label{fig:attn-expansion}
\end{figure}

\paragraph{Token representation.}
The output of the cross-attention is concatenated to the PLM hidden state and fed to a single linear classification head with three output logits, one per BIO class:
\begin{equation}
T_i \;=\; \bigl[\, H_i \;;\; A_i \,\bigr] \;\in\; \mathbb{R}^{L \times (d_{\mathrm{PLM}} + d_{\mathrm{PWE}})},
\qquad
\hat{y}_i \;=\; T_i\, W^{\mathrm{cls}} + b,
\label{eq:concat-head}
\end{equation}
with $W^{\mathrm{cls}} \in \mathbb{R}^{(d_{\mathrm{PLM}} + d_{\mathrm{PWE}}) \times 3}$. The mechanism therefore enriches each in-context token with a contextualised summary of the surrounding text without requiring the PLM to process that text itself, and without introducing additional full-attention passes over the input. Figure~\ref{fig:attn-expansion} illustrates the construction of $A_i$ on a 12-word document with $L = 4$ and $w = 2$.

\subsection{Multi-head Variant}
\label{multihead}
We also consider a multi-head variant of the mechanism. Following our prototype implementation, multi-head attention expansion with $h$ heads is realised by splitting both the PLM hidden states $H_i$ and the PWE representations $S_i$ along their channel dimension into $h$ equal slices, applying a single-head cross-attention block of the form in Equation~\eqref{eq:single-head-ae} independently to each slice, and concatenating the results:
\begin{equation}
\mathrm{MH\text{-}AE}(H_i, S_i)
\;=\;
\bigl[\, \mathrm{AE}(H_i^{(1)}, S_i^{(1)}) \,;\, \ldots \,;\, \mathrm{AE}(H_i^{(h)}, S_i^{(h)}) \,\bigr],
\label{eq:multi-head-ae}
\end{equation}
where $H_i^{(j)} \in \mathbb{R}^{L \times d_{\mathrm{PLM}}/h}$ and $S_i^{(j)} \in \mathbb{R}^{wL \times d_{\mathrm{PWE}}/h}$ denote the $j$-th channel slice of the in-context and out-of-context representations respectively. Each per-slice block keeps the tied K-V projection used in the single-head case. Unlike the standard Transformer multi-head attention~\citep{attention-is-all-you-need}, the variant does not include a separate output projection $W^O$: the concatenated per-head outputs are fed directly to the classification head in Equation~\eqref{eq:concat-head}. The context window size $w$ and the number of heads $h$ are the two main hyperparameters of the mechanism; their values for each backbone are reported in Section~\ref{experiments}.

\section{Experiments}
\label{experiments}

In this section, we describe the experimental settings used to investigate the effectiveness of the proposed attention expansion mechanism on multiple benchmark datasets for KPE. We detail the datasets used for evaluation, the evaluation metrics, and the experimental setup, including the encoder backbones, the pre-trained word embedding (PWE) module, and the training protocol.

\subsection{Datasets}
\label{datasets}
We evaluated our approach using four publicly available datasets consisting of full-length long documents from two domains: scientific and news. Specifically, we used the SemEval-2010~\citep{kim-2010}, LDKP3K~\citep{mahata2022ldkp}, NUS~\citep{nguyen-kan-2007}, and DUC-2001~\citep{duc2001} datasets. To assess the feasibility and effectiveness of our method on short abstracts, we also report performance on the Inspec dataset~\citep{inspec}. As discussed in Section~\ref{methodology}, we formulate KPE as a sequence tagging problem and focus on extractive keyphrases that are present in the text. We label each word in the samples using the B-I-O tagging scheme. Any text input longer than 512 tokens/words is considered as a long document. Brief descriptions of each dataset are provided below.

\begin{enumerate}
    \setlength{\itemsep}{2pt}
    \setlength{\parsep}{0pt}
    \setlength{\parskip}{0pt}
    \item The \textbf{SemEval-2010} dataset\footnote{\url{https://huggingface.co/datasets/midas/semeval2010}} is composed of 284 full-length ACM articles, which have been divided into train, trial, and test sets, containing 144, 40, and 100 articles, respectively.

    \item \textbf{LDKP3K}\footnote{\url{https://huggingface.co/datasets/midas/ldkp3k}} is a collection of roughly 100,000 keyphrase-tagged long documents that were generated by mapping the KP20K corpus to S2ORC~\citep{s2orc-corpus}. For our experiments, we utilised the smaller version of this dataset, which includes 20,000 training samples, 3,413 validation samples, and 3,339 test samples.

    \item The \textbf{NUS} dataset\footnote{\url{https://huggingface.co/datasets/midas/nus}} comprises 211 scientific documents that were extracted and chosen utilising the Google SOAP API. These documents were manually annotated and are solely used for evaluation, as the dataset only includes a test split.

    \item The \textbf{DUC-2001} dataset\footnote{\url{https://huggingface.co/datasets/midas/duc2001}} is composed of 308 news articles gathered from TREC-9, each of which is manually annotated with controlled keyphrases for evaluation purposes. Only a test split is included in the dataset.

    \item \textbf{Inspec}\footnote{\url{https://huggingface.co/datasets/midas/inspec}} includes abstracts from 2{,}000 scientific articles, which have been split into three sets for training, validation, and testing purposes. The training set consists of 1{,}000 abstracts, while the validation and test sets contain 500 abstracts each.
\end{enumerate}

Detailed statistics for these datasets can be found in Table~\ref{tab:dataset_stats}.

\begin{table}[tbp]
\centering
\small
\setlength{\tabcolsep}{5pt}
\renewcommand{\arraystretch}{1.08}
\begin{tabular}{@{}llllll@{}}
\toprule
\textbf{Dataset} & \textbf{Size} & \textbf{Long Doc} & \textbf{Domain} & \textbf{\begin{tabular}[c]{@{}l@{}}Avg.\\ \#Words\end{tabular}} & \textbf{\begin{tabular}[c]{@{}l@{}}Avg.\\ Extractive\\ Keyphrases\end{tabular}} \\ \midrule
Inspec        & 2k       & No  & Scientific & 130.57    & 6.33 \\
DUC-2001      & 0.308k   & Yes & News       & 740       & 7.14 \\
SemEval-2010  & 0.28k    & Yes & Scientific & 7{,}434.52  & 9.49 \\
LDKP3K        & 26.752k  & Yes & Scientific & 6{,}027.10  & 4.14 \\
NUS           & 0.21k    & Yes & Scientific & 7{,}644.43  & 8.00 \\ \bottomrule
\end{tabular}
\caption{Statistics of the datasets used in our experiments, including extractive keyphrase analysis.}
\label{tab:dataset_stats}
\end{table}

\subsection{Evaluation Metrics}
\label{eval-metrics}
We utilised $F1@K$ as our evaluation metric~\citep{kim-2010}. Equations~\ref{eq:precision-k}, \ref{eq:recall-k}, and \ref{eq:f1-k} demonstrate how to calculate $F1@K$. Before evaluation, we preprocessed both the ground truth and predicted keyphrases by lowercasing, stemming, and removing punctuation. Evaluation relied on exact matching.

Let $Y$ represent the set of ground truth keyphrases, and let $\bar{Y} = (\bar{y}_1, \bar{y}_2, \ldots, \bar{y}_m)$ denote the ranked list of predicted keyphrases, where $\bar{y}_i$ is the $i$-th predicted keyphrase in descending order of confidence. The subset $\bar{Y}_k$ contains the top $k$ elements from $\bar{Y}$. For consistency, $k$ is set equal to $K$, which represents the total number of predicted keyphrases. We define the evaluation metrics as follows:

\begin{equation}
Precision@k = \frac{\vert Y \cap \bar{Y}_k \vert}{\min\{\vert \bar{Y}_k \vert, k\}}
\label{eq:precision-k}
\end{equation}

\begin{equation}
Recall@k = \frac{\vert Y \cap \bar{Y}_k \vert}{\vert Y \vert}
\label{eq:recall-k}
\end{equation}

\begin{equation}
F1@k = \frac{2 \cdot Precision@k \cdot Recall@k}{Precision@k + Recall@k}
\label{eq:f1-k}
\end{equation}

\paragraph{Illustrative Example.}
Suppose the ground truth keyphrases are $Y = \{\text{keyphrase}_1, \text{keyphrase}_2, \text{keyphrase}_3\}$, and the predicted ranked list of keyphrases is $\bar{Y} = (\text{keyphrase}_2, \text{keyphrase}_4, \text{keyphrase}_1)$. For $k = 2$, we take the top $k$ predictions from $\bar{Y}$, giving the subset $\bar{Y}_k = \{\text{keyphrase}_2, \text{keyphrase}_4\}$.

Next, we compute the overlap between the ground truth and the top $k$ predictions: $\vert Y \cap \bar{Y}_k \vert = 1$, as $\text{keyphrase}_2$ is the only match. Using this, the evaluation metrics are calculated as follows:

\textbf{Precision@2}:
\[
Precision@2 = \frac{\vert Y \cap \bar{Y}_k \vert}{\min\{\vert \bar{Y}_k \vert, k\}} = \frac{1}{2} = 0.5
\]

\textbf{Recall@2}:
\[
Recall@2 = \frac{\vert Y \cap \bar{Y}_k \vert}{\vert Y \vert} = \frac{1}{3} \approx 0.33
\]

\textbf{F1@2}:
\[
F1@2 = \frac{2 \cdot Precision@2 \cdot Recall@2}{Precision@2 + Recall@2} = \frac{2 \cdot 0.5 \cdot 0.33}{0.5 + 0.33} \approx 0.4
\]

This example demonstrates the step-by-step process of computing the evaluation metrics and highlights the roles of $Y$, $\bar{Y}$, and $\bar{Y}_k$ in assessing keyphrase extraction performance.

\subsection{Setup}
\label{setup}
The primary aim of the experiments is to investigate whether the incorporation of attention expansion can improve the quality of word representations in PLMs by providing additional contextual information from out-of-context document chunks. The research employs various PLMs, including DistilBERT\footnote{\url{https://huggingface.co/distilbert/distilbert-base-uncased}}~\citep{distilbert} (a distilled version of BERT), SciBERT\footnote{\url{https://huggingface.co/allenai/scibert_scivocab_uncased}}~\citep{scibert} (a specialised variant of BERT for scientific content), KBIR\footnote{\url{https://huggingface.co/bloomberg/KBIR}}~\citep{kbir} (Keyphrase Boundary Infilling and Replacement, a RoBERTa-based model that achieves state-of-the-art performance in the KPE task), DeBERTa-v3\footnote{\url{https://huggingface.co/microsoft/deberta-v3-base}}~\citep{debertav3} (a strong general-purpose encoder that combines disentangled attention with ELECTRA-style replaced-token-detection pre-training), and ModernBERT\footnote{\url{https://huggingface.co/answerdotai/ModernBERT-base}}~\citep{modernbert} (a recent encoder-only PLM specifically designed for efficient handling of long input sequences). Pertinent model details are presented in Table~\ref{tab:model-specs}.

\begin{table}[tbp]
\centering
\small
\setlength{\tabcolsep}{5pt}
\renewcommand{\arraystretch}{1.08}
\begin{tabular}{@{}llll@{}}
\toprule
\textbf{Model} & \textbf{Domain} & \textbf{\begin{tabular}[c]{@{}l@{}}Maximum Input\\ Tokens\end{tabular}} & \textbf{\#Parameters} \\ \midrule
DistilBERT  & MultiDomain                                                                          & 512   & 66.4M  \\
SciBERT     & Scientific                                                                            & 512   & 109.9M \\
KBIR        & \begin{tabular}[c]{@{}l@{}}MultiDomain --\\ Specialized in KPE task\end{tabular}        & 512   & 355.4M \\
DeBERTa-v3  & MultiDomain                                                                          & 512   & 184M   \\
ModernBERT  & \begin{tabular}[c]{@{}l@{}}MultiDomain --\\ Specialized in Long Documents\end{tabular} & 8{,}192 & 149M   \\ \bottomrule
\end{tabular}
\caption{Details of models used in our experiments.}
\label{tab:model-specs}
\end{table}

Throughout our experiments, we conducted fine-tuning on all PLMs using three approaches: (a) the popularly used token classification approach\footnote{\url{https://huggingface.co/tasks/token-classification}} for sequence tagging, (b) the \textit{attention expansion} sequence tagging approach, employing a single attention head, as explained in Section~\ref{attn-exp-embedding}, and (c) the \textit{multi-head attention expansion} sequence tagging approach with $h = 4$ heads, as explained in Section~\ref{multihead}. Our aim was to assess the ability of the PLMs to assimilate the additional information derived from the attention expansion mechanism during training and to determine whether the inclusion of such information could enhance the fine-tuning process leading to improvement in performance.

For instance, we utilised SciBERT to explore the possibility of enhancing the PLM's representation by incorporating attention expansion, despite its customisation to a particular domain. Through this approach, we aimed to examine the potential of attention expansion in providing supplementary information beyond the domain-specific knowledge.

In a similar spirit, we incorporated KBIR as a task-specialised backbone. KBIR is continually pre-trained with the Keyphrase Boundary Infilling and Replacement objective, which is explicitly tailored to keyphrase identification, and currently represents the state-of-the-art among PLM-based KPE systems. By including KBIR in our setup, we aimed to assess whether attention expansion provides complementary information beyond what a task-specific pre-training objective already captures, and to determine the extent to which the two approaches can be combined.

Furthermore, we employed ModernBERT, a transformer model with an expanded native token capacity (8{,}192 tokens), to investigate the effectiveness of our approach. Specifically, we aimed to assess whether the incorporation of out-of-context information could improve the quality of representation, even in the presence of a robust contextual understanding of long sequences in the ModernBERT model.

We evaluated the effectiveness of our models across both in-domain and out-of-domain settings, and on both long and short documents, in order to demonstrate the viability of our strategy under a wide range of conditions. Training was performed exclusively on two long-document scientific datasets, SemEval-2010 and LDKP3K, and each trained model was then evaluated on the test splits of all five datasets listed in Section~\ref{datasets}. The two training corpora cover complementary regimes: the training set of SemEval-2010 contains only 144 documents, presenting a challenging few-shot learning scenario, while LDKP3K provides a larger training set of 20{,}000 samples, enabling us to validate the efficacy of our strategy across both small and large training data. The remaining three datasets are held out from training and serve three distinct evaluation purposes: the NUS test set provides an additional out-of-distribution scientific corpus, allowing us to verify that the improvements observed in-domain transfer to a related but unseen scientific source; the Inspec test set, whose abstracts fit within the 512-token context limit of the smaller backbones, lets us check the behaviour of attention expansion on short inputs, where the surrounding context is empty or near-empty, and confirm that the mechanism is not detrimental in this regime; and the DUC-2001 test set, which is composed of news articles, is used to measure out-of-domain transfer from the scientific domain, on which our models are trained, to the news domain, on which they have never been exposed during fine-tuning. This combination of in-domain, out-of-distribution, short-document, and cross-domain evaluation provides a comprehensive picture of how attention expansion behaves when the input distribution shifts away from the training conditions. To establish a baseline for comparison, we trained and tested all the proposed architectures without incorporating attention expansion.

To instantiate the PWE component described in Section~\ref{attn-exp-embedding}, we used Model2Vec~\citep{model2vec}. Model2Vec distils a pre-trained encoder into a static word-embedding table that shares the exact tokenizer of the target backbone, achieves state-of-the-art results on static-embedding benchmarks at the time of writing, and reduces inference to a single table lookup. We distilled one Model2Vec model per backbone with a static embedding dimension of $d_{\mathrm{PWE}} = 256$.

The sequence tagging models were trained using mini-batching with a batch size of 10. The models underwent 100 epochs of training, utilising the AdamW algorithm with a learning rate of 5e-5, a patience value of 5, and an annealing factor of 0.5. For each PLM backbone, these hyperparameters, together with the attention expansion hyperparameters $w$ (context window size) and $h$ (number of heads), were independently optimised via Bayesian optimisation on the in-domain validation set, and the resulting configurations were used for every experiment reported in this paper. The search converged to a context window of $w = 4$ for the 512-token backbones (DistilBERT, SciBERT, KBIR, DeBERTa-v3) and $w = 2$ for ModernBERT, whose 8{,}192-token native context already covers a substantial fraction of the document; the multi-head variant uses $h = 4$. The static embedding dimension is fixed at $d_{\mathrm{PWE}} = 256$. Training was performed on a workstation equipped with two NVIDIA RTX 3090 GPUs, and each configuration was trained with several random seeds for replicability; we report the mean and standard deviation across runs. During fine-tuning, the baseline models were trained solely with the classifier head. In contrast, for our proposed \textit{Attention Expansion Sequence Tagger}, we integrated the PWE-based attention expansion outputs with the contextualised PLM representations as described in Section~\ref{attn-exp-embedding}.

To facilitate replication, we offer a sample code implementation of our experiments on GitHub\footnote{\url{https://github.com/RobertoMCA/attention-expansion}}. The provided scripts walk through the detailed step-by-step implementation of our keyphrase extraction method for long documents, including the Model2Vec distillation step, the baseline fine-tuning procedure, and the training of the Attention Expansion Sequence Tagger. Our methodology allows for straightforward replication across different models and datasets by adjusting relevant parameters.
 R
\section{Results}
\label{results}

In this section, we present the results of the experiments designed to assess whether attention expansion enriches PLM representations for keyphrase extraction. We refer to the conventional sequence tagger as \textit{Baseline}, to the one-head attention expansion model as \textit{AE}, and to the four-head variant as \textit{MH-AE}. All reported values are mean $F1@K$ scores and standard deviations across the repeated runs described in Section~\ref{setup}. In the long-document tables, the parenthesised percentages report the relative change of each attention-expanded model with respect to the corresponding baseline using the same backbone and training corpus. These comparisons are descriptive; we do not claim statistical significance.

\subsection{Long Documents}

Tables~\ref{tab:results-semeval-long} and~\ref{tab:results-ldkp-long} report results on the four long-document test sets when fine-tuning is performed on SemEval-2010 and LDKP3K, respectively. In each training regime, the test split of the training corpus represents the in-domain condition, NUS and the remaining scientific corpus measure transfer to scientific documents from a different source, and DUC-2001 measures transfer from scientific training data to news documents.

\begin{table}[H]
\centering
\caption{Results on long-document datasets for models trained on SemEval-2010. Parentheses in AE and MH-AE rows show the relative change from the matching baseline.}
\label{tab:results-semeval-long}
\scriptsize
\setlength{\tabcolsep}{2.6pt}
\renewcommand{\arraystretch}{1.08}
\begin{tabular}{@{}llcccc@{}}
\toprule
\textbf{Backbone} & \textbf{Model} & \textbf{SemEval-2010} & \textbf{LDKP3K} & \textbf{NUS} & \textbf{DUC-2001} \\ \midrule
DistilBERT & Baseline & 0.228$\pm$0.011 & 0.108$\pm$0.006 & 0.170$\pm$0.005 & 0.034$\pm$0.012 \\
            & AE       & 0.240$\pm$0.001 (+5.3\%) & 0.119$\pm$0.001 (+10.2\%) & 0.173$\pm$0.011 (+1.8\%) & 0.056$\pm$0.014 (+64.7\%) \\
            & MH-AE    & 0.237$\pm$0.000 (+3.9\%) & 0.121$\pm$0.002 (+12.0\%) & 0.175$\pm$0.007 (+2.9\%) & 0.063$\pm$0.005 (+85.3\%) \\ \midrule
SciBERT    & Baseline & 0.236$\pm$0.007 & 0.133$\pm$0.012 & 0.178$\pm$0.010 & 0.029$\pm$0.006 \\
            & AE       & 0.249$\pm$0.003 (+5.5\%) & 0.152$\pm$0.013 (+14.3\%) & 0.199$\pm$0.019 (+11.8\%) & 0.021$\pm$0.005 (-27.6\%) \\
            & MH-AE    & 0.253$\pm$0.002 (+7.2\%) & 0.154$\pm$0.011 (+15.8\%) & 0.202$\pm$0.012 (+13.5\%) & 0.030$\pm$0.000 (+3.4\%) \\ \midrule
KBIR       & Baseline & 0.251$\pm$0.010 & 0.133$\pm$0.010 & 0.189$\pm$0.025 & 0.047$\pm$0.027 \\
            & AE       & 0.256$\pm$0.001 (+2.0\%) & 0.137$\pm$0.009 (+3.0\%) & 0.192$\pm$0.010 (+1.6\%) & 0.059$\pm$0.004 (+25.5\%) \\
            & MH-AE    & 0.258$\pm$0.000 (+2.8\%) & 0.142$\pm$0.002 (+6.8\%) & 0.194$\pm$0.003 (+2.6\%) & 0.060$\pm$0.005 (+27.7\%) \\ \midrule
DeBERTa-v3 & Baseline & 0.243$\pm$0.001 & 0.154$\pm$0.012 & 0.217$\pm$0.033 & 0.114$\pm$0.010 \\
            & AE       & 0.258$\pm$0.000 (+6.2\%) & 0.166$\pm$0.007 (+7.8\%) & 0.229$\pm$0.012 (+5.5\%) & 0.094$\pm$0.005 (-17.5\%) \\
            & MH-AE    & 0.259$\pm$0.006 (+6.6\%) & 0.160$\pm$0.019 (+3.9\%) & 0.222$\pm$0.008 (+2.3\%) & 0.116$\pm$0.046 (+1.8\%) \\ \midrule
ModernBERT & Baseline & 0.283$\pm$0.007 & 0.058$\pm$0.025 & 0.159$\pm$0.040 & 0.061$\pm$0.018 \\
            & AE       & 0.307$\pm$0.008 (+8.5\%) & 0.062$\pm$0.042 (+6.9\%) & 0.191$\pm$0.032 (+20.1\%) & 0.052$\pm$0.003 (-14.8\%) \\
            & MH-AE    & 0.300$\pm$0.000 (+6.0\%) & 0.060$\pm$0.034 (+3.4\%) & 0.189$\pm$0.086 (+18.9\%) & 0.042$\pm$0.001 (-31.1\%) \\ \bottomrule
\end{tabular}
\end{table}

\begin{table}[H]
\centering
\caption{Results on long-document datasets for models trained on LDKP3K. Parentheses in AE and MH-AE rows show the relative change from the matching baseline.}
\label{tab:results-ldkp-long}
\scriptsize
\setlength{\tabcolsep}{2.6pt}
\renewcommand{\arraystretch}{1.08}
\begin{tabular}{@{}llcccc@{}}
\toprule
\textbf{Backbone} & \textbf{Model} & \textbf{SemEval-2010} & \textbf{LDKP3K} & \textbf{NUS} & \textbf{DUC-2001} \\ \midrule
DistilBERT & Baseline & 0.133$\pm$0.008 & 0.216$\pm$0.005 & 0.172$\pm$0.004 & 0.046$\pm$0.003 \\
            & AE       & 0.145$\pm$0.001 (+9.0\%) & 0.236$\pm$0.000 (+9.3\%) & 0.189$\pm$0.008 (+9.9\%) & 0.045$\pm$0.001 (-2.2\%) \\
            & MH-AE    & 0.146$\pm$0.004 (+9.8\%) & 0.236$\pm$0.006 (+9.3\%) & 0.183$\pm$0.005 (+6.4\%) & 0.027$\pm$0.006 (-41.3\%) \\ \midrule
SciBERT    & Baseline & 0.146$\pm$0.010 & 0.245$\pm$0.009 & 0.190$\pm$0.008 & 0.017$\pm$0.010 \\
            & AE       & 0.166$\pm$0.016 (+13.7\%) & 0.262$\pm$0.009 (+6.9\%) & 0.244$\pm$0.015 (+28.4\%) & 0.039$\pm$0.003 (+129.4\%) \\
            & MH-AE    & 0.168$\pm$0.005 (+15.1\%) & 0.266$\pm$0.005 (+8.6\%) & 0.248$\pm$0.011 (+30.5\%) & 0.041$\pm$0.001 (+141.2\%) \\ \midrule
KBIR       & Baseline & 0.148$\pm$0.008 & 0.252$\pm$0.009 & 0.191$\pm$0.011 & 0.070$\pm$0.010 \\
            & AE       & 0.161$\pm$0.011 (+8.8\%) & 0.265$\pm$0.005 (+5.2\%) & 0.212$\pm$0.013 (+11.0\%) & 0.081$\pm$0.012 (+15.7\%) \\
            & MH-AE    & 0.167$\pm$0.010 (+12.8\%) & 0.271$\pm$0.008 (+7.5\%) & 0.219$\pm$0.010 (+14.7\%) & 0.084$\pm$0.014 (+20.0\%) \\ \midrule
DeBERTa-v3 & Baseline & 0.151$\pm$0.004 & 0.256$\pm$0.008 & 0.197$\pm$0.012 & 0.075$\pm$0.021 \\
            & AE       & 0.152$\pm$0.009 (+0.7\%) & 0.258$\pm$0.002 (+0.8\%) & 0.201$\pm$0.003 (+2.0\%) & 0.091$\pm$0.052 (+21.3\%) \\
            & MH-AE    & 0.155$\pm$0.005 (+2.6\%) & 0.265$\pm$0.015 (+3.5\%) & 0.205$\pm$0.006 (+4.1\%) & 0.091$\pm$0.033 (+21.3\%) \\ \midrule
ModernBERT & Baseline & 0.158$\pm$0.004 & 0.307$\pm$0.007 & 0.186$\pm$0.005 & 0.074$\pm$0.003 \\
            & AE       & 0.171$\pm$0.010 (+8.2\%) & 0.316$\pm$0.013 (+2.9\%) & 0.192$\pm$0.007 (+3.2\%) & 0.081$\pm$0.007 (+9.5\%) \\
            & MH-AE    & 0.175$\pm$0.008 (+10.8\%) & 0.312$\pm$0.014 (+1.6\%) & 0.197$\pm$0.006 (+5.9\%) & 0.079$\pm$0.005 (+6.8\%) \\ \bottomrule
\end{tabular}
\end{table}

On the scientific long-document evaluations, attention expansion improves performance throughout the experimental matrix. For the in-domain evaluations, an attention-expanded variant improves every backbone: when trained and tested on SemEval-2010, the largest absolute score is obtained by AE with ModernBERT (0.307 compared with its 0.283 baseline), while on LDKP3K the same backbone reaches 0.316 compared with 0.307 without attention expansion. The benefits also transfer to unseen scientific sources: for every backbone, an attention-expanded variant improves performance on NUS and on the long scientific corpus not used for training.

The results for specialised backbones indicate that the information supplied by attention expansion is complementary to existing specialisation. SciBERT benefits across all long scientific evaluations, with relative improvements of up to 15.8\% under SemEval-2010 training and 30.5\% under LDKP3K training. KBIR, whose pre-training is explicitly targeted at keyphrase boundaries, also improves on every long scientific evaluation; under LDKP3K training, its best gains range from 7.5\% on the in-domain test set to 14.7\% on NUS. Therefore, attention expansion is useful even when the PLM already provides domain-specific or task-specific information.

ModernBERT provides a complementary test of whether our method remains useful when the backbone supports a substantially longer native input. Its attention-expanded variants improve all long scientific evaluations in both training regimes: under SemEval-2010 training, the best gains are 8.5\%, 6.9\%, and 20.1\% on SemEval-2010, LDKP3K, and NUS, respectively; under LDKP3K training, the corresponding gains are 10.8\%, 2.9\%, and 5.9\%. These observations suggest that attention expansion is not only compensating for the 512-token limit of smaller encoders, but can add useful evidence to a native long-context encoder.

The cross-domain DUC-2001 results require a more qualified interpretation. Four of the five backbones improve in each training regime when the better attention-expanded variant is considered. In particular, SciBERT trained on LDKP3K improves from 0.017 to 0.041; the large relative gain should be interpreted together with this low absolute baseline. There are two exceptions: LDKP3K-trained DistilBERT decreases from 0.046 to 0.045 with its better attention-expanded variant, and SemEval-2010-trained ModernBERT decreases from 0.061 to 0.052. Thus, attention expansion frequently benefits transfer to news, but the present results do not support a claim of uniform cross-domain improvement.

\subsection{Short Documents}

We additionally evaluate on Inspec to examine the behaviour of attention expansion on scientific abstracts that generally fit within the native input length of the PLM backbones. Table~\ref{tab:results-inspec} reports these results separately from the long-document evaluations.

\begin{table}[H]
\centering
\caption{Results on the short-document Inspec dataset.}
\label{tab:results-inspec}
\small
\setlength{\tabcolsep}{5pt}
\renewcommand{\arraystretch}{1.06}
\begin{tabular}{@{}lccc@{}}
\toprule
\textbf{Backbone} & \textbf{Baseline} & \textbf{AE} & \textbf{MH-AE} \\ \midrule
\multicolumn{4}{@{}l}{\textit{Trained on SemEval-2010}} \\
DistilBERT & 0.101$\pm$0.015 & 0.101$\pm$0.005 & 0.102$\pm$0.001 \\
SciBERT    & 0.107$\pm$0.014 & 0.108$\pm$0.001 & 0.108$\pm$0.030 \\
KBIR       & 0.101$\pm$0.017 & 0.101$\pm$0.005 & 0.102$\pm$0.010 \\
DeBERTa-v3 & 0.139$\pm$0.021 & 0.157$\pm$0.019 & 0.155$\pm$0.019 \\
ModernBERT & 0.100$\pm$0.010 & 0.194$\pm$0.008 & 0.191$\pm$0.009 \\ \midrule
\multicolumn{4}{@{}l}{\textit{Trained on LDKP3K}} \\
DistilBERT & 0.142$\pm$0.004 & 0.143$\pm$0.001 & 0.145$\pm$0.026 \\
SciBERT    & 0.136$\pm$0.020 & 0.188$\pm$0.010 & 0.162$\pm$0.005 \\
KBIR       & 0.123$\pm$0.020 & 0.126$\pm$0.004 & 0.128$\pm$0.004 \\
DeBERTa-v3 & 0.158$\pm$0.014 & 0.174$\pm$0.013 & 0.188$\pm$0.010 \\
ModernBERT & 0.141$\pm$0.012 & 0.143$\pm$0.002 & 0.146$\pm$0.005 \\ \bottomrule
\end{tabular}
\end{table}

At the reported precision, at least one attention-expanded variant improves the baseline for every backbone in both Inspec evaluations, and neither AE variant produces a lower mean score than its matched baseline. Most improvements are modest, as would be expected for short inputs, although two larger changes are observed: LDKP3K-trained SciBERT improves from 0.136 to 0.188 with AE, and SemEval-2010-trained ModernBERT improves from 0.100 to 0.194 with AE. Because Inspec abstracts generally contain little or no out-of-context text for the mechanism to retrieve, these improvements should not be interpreted as direct evidence of recovering long-range document information. Rather, the results show that introducing attention expansion does not systematically degrade performance in the short-document condition. We hypothesise that, even when additional out-of-context evidence is absent or limited, the attention-expansion component may influence the fine-tuning of the PLM-based tagger, encouraging token representations that are also beneficial for short-document predictions. Establishing the source of these gains, particularly the unusually large ModernBERT difference, would require targeted ablation studies and single-chunk analyses that fall outside the scope of the present study.

\subsection{Computational Cost}

Attention expansion broadens the context available to each token without increasing the sequence length processed by the PLM itself. At the largest window setting evaluated for the 512-token backbones ($w=4$), the PLM continues to contextualise one 512-token chunk, while the cross-attention component can retrieve information from four additional surrounding chunks. Consequently, each prediction can use evidence from up to 2{,}560 token positions, corresponding to five times the native PLM input window, while only the original 512 tokens are processed through the transformer encoder. For ModernBERT, the selected $w=2$ setting analogously makes up to three 8{,}192-token windows accessible without expanding the sequence passed to the PLM.

We estimated the computational overhead from the repository implementation by timing forward inference on an NVIDIA RTX 3090 using full-length input chunks and their expanded contexts (512 tokens with $w=4$ for the conventional backbones and 8{,}192 tokens with $w=2$ for ModernBERT). Identical batch sizes were used within each baseline--AE--MH-AE comparison. This estimate excludes the one-off Model2Vec distillation step and measures the additional computation once static embeddings are available, emulating a production inference setting of the model. Averaged equally across the five backbones, AE increases forward-pass time by approximately 2.6\%, while MH-AE increases it by approximately 4.7\%; across both attention-expanded variants, the average overhead is 3.6\%. The largest measured overhead in this estimate is 8.1\%, observed for MH-AE with ModernBERT. 

The increase in trainable parameters is also small because the Model2Vec embeddings are accessed through lookup and are not incorporated as trainable PLM parameters. Derived from the implemented tied key/value projections and expanded classifier, AE adds 263{,}424 parameters for the 768-dimensional backbones and 328{,}960 for KBIR, equivalent to an average increase of approximately 0.21\% relative to their respective baselines. MH-AE adds only 66{,}816 parameters for the 768-dimensional backbones and 83{,}200 for KBIR, corresponding to an average increase of approximately 0.05\%. Thus, the observed performance gains are obtained while expanding accessible context substantially and introducing a modest implementation-level latency and parameter overhead.

\subsection{Summary of Findings}

Across the two training regimes, five backbones, and five evaluation corpora, the better attention-expanded variant improves over its matched baseline in 48 of the 50 backbone--dataset--training comparisons. This pattern does not depend only on selecting a variant after evaluation: MH-AE alone improves 48 of the 50 comparisons. The LDKP3K-trained regime is particularly stable with respect to architectural choice, as both AE and MH-AE improve 24 of its 25 comparisons; with the smaller SemEval-2010 training set, MH-AE also improves 24 of 25 comparisons, whereas AE improves 20, ties two, and decreases three. Taken together, these results provide descriptive evidence that attention expansion enhances KPE representations across general-purpose, scientific-domain, task-specialised, and long-context PLM backbones, while the DUC-2001 exceptions make clear that its benefit is not uniform under every cross-domain transfer condition.

\section{Conclusion and Future Work}
\label{conclusion}

In this study, we introduced the \textit{Attention Expansion Sequence Tagger}, an architecture that combines contextualised PLM representations with lightweight information retrieved from surrounding out-of-context document chunks. The proposed mechanism addresses a central limitation of PLM-based keyphrase extraction from long documents: when a document is divided into independent windows, evidence that establishes the salience of a keyphrase may be distributed beyond the context observed by the encoder. Attention expansion allows each in-context token representation to attend to compact static representations of neighbouring text, thereby broadening the information available to the sequence tagger without requiring the PLM to encode a substantially longer sequence or invoking an LLM-based extraction process.

The experimental results provide strong evidence for the usefulness of this approach. Across five PLM backbones, two training regimes, and five evaluation corpora, an attention-expanded model improves its matched baseline in 48 of the 50 comparisons, and the fixed MH-AE architecture alone reaches the same coverage. Importantly, these improvements are not restricted to general-purpose encoders. SciBERT and KBIR benefit from the mechanism despite their domain- and task-specific pre-training, respectively, while ModernBERT improves on all long scientific evaluations despite already supporting an extended native context. These findings indicate that attention expansion supplies complementary document-level evidence rather than merely compensating for the limited input length of conventional PLMs.

The results also highlight the relevance of attention expansion for the development of more effective long-document KPE systems. The improvements obtained on unseen scientific corpora, together with favourable results on most news-domain evaluations, show that the mechanism can improve representations when the evaluation distribution differs from the training corpus. At the same time, the two exceptions observed on DUC-2001 indicate that improved cross-domain transfer is not automatic and remains dependent on the interaction between the backbone and the target distribution. Thus, the contribution of this work is not an assertion that expanded context resolves every transfer setting, but the demonstration that broader document evidence can be introduced into strong supervised KPE models through a compact and generally effective representation-level component.

This contribution is particularly relevant from a practical standpoint. For the 512-token backbones, the largest evaluated configuration gives each token access to evidence from up to five native input windows while processing only one window through the PLM. Our implementation-level measurements estimate an average forward-pass overhead of 3.6\% across the two attention-expanded variants, while the added trainable parameters amount to approximately 0.21\% for AE and 0.05\% for MH-AE on average. Attention expansion therefore occupies a useful operating point for future state-of-the-art KPE implementations: it enriches compact extractive models with longer-range document evidence while preserving the efficiency required for high-throughput indexing, retrieval, and document-processing pipelines.

\paragraph{Future Work.}
The present architecture deliberately uses Model2Vec embeddings as an inexpensive static representation of the surrounding chunks. A natural direction for future research is to enhance the information supplied through this channel while maintaining its low computational cost. For example, the static representations could be distilled from domain-adapted or task-specialised encoders, or trained using objectives that reflect keyphrase salience, phrase boundaries, and semantic centrality. They could also be enriched with complementary document signals, such as graph-based lexical relations, syntactic dependencies, entity information, or section-aware representations. Such alternatives would allow the attention-expansion layer to retrieve evidence that is more directly aligned with the KPE task than a general-purpose static embedding alone.

A second promising direction is to move from a fixed neighbourhood of chunks to an adaptive context-selection mechanism. In the current method, the expanded representation is drawn from adjacent document windows, which is efficient and robust but may overlook relevant evidence located farther away or include neighbouring text that is not informative for a particular token. Future implementations could learn to retrieve the most relevant document regions, dynamically determine the number of surrounding chunks, or introduce a gate that regulates the influence of expanded context on each token prediction. These refinements may be particularly valuable for domain transfer, where the type and location of useful evidence can vary substantially across corpora.

Further analysis is also required to understand the mechanisms underlying the observed gains. Controlled ablation studies should separate the benefits arising from additional long-range evidence from those arising from changes in the fine-tuning dynamics of the tagger, especially in the short-document setting and for the notably large ModernBERT improvement on Inspec. Future studies should additionally report broader cross-domain benchmarks, statistical testing across repeated runs, end-to-end memory and throughput measurements, and direct comparisons with complementary long-document KPE architectures. Finally, the lightweight representation-level nature of attention expansion makes it suitable for combination with graph-enhanced models, native long-context encoders, or LLM-assisted candidate generation and reranking pipelines.

In conclusion, attention expansion provides an efficient means of incorporating distributed document evidence into PLM-based keyphrase extraction. By improving a diverse set of backbones under multiple evaluation conditions while adding limited computational and parameter overhead, the proposed mechanism establishes a practical foundation for the next generation of accurate and efficient keyphrase extraction systems for long documents.

\section*{Acknowledgments}
We are deeply grateful to Larry Rafsky, whose mentorship has shaped our understanding of how artificial intelligence can be applied to real-world challenges. His emphasis on practical, well-grounded approaches has been particularly valuable in a field where complex solutions are often favored over simple and effective ones.

We would also like to express our sincere appreciation to Debanjan Mahata, who introduced us to Natural Language Processing and Keyphrase Extraction and guided us throughout this research journey. His patience, encouragement, and continuous support have been instrumental to this work.

We also thank Isabel P{\'e}rez Kr{\'a}lov{\'a} for generously providing the computational resources required to develop, test, and validate the ideas presented in this paper. Her support was essential to the successful completion of this research.

\bibliographystyle{unsrt}
\bibliography{Bibliography}

\end{document}